\documentclass[letterpaper]{article} 
\usepackage{aaai2026}  
\usepackage{times}  
\usepackage{helvet}  
\usepackage{courier}  
\usepackage[hyphens]{url}  
\usepackage{graphicx} 
\usepackage{natbib}  
\usepackage{caption} 
\usepackage{algorithm}
\usepackage{algorithmic}

\usepackage{newfloat}
\usepackage{listings}
\DeclareCaptionStyle{ruled}{labelfont=normalfont,labelsep=colon,strut=off} 
\floatstyle{ruled}
\newfloat{listing}{tb}{lst}{}
\floatname{listing}{Listing}
\usepackage{paralist} 
\usepackage{booktabs} 
\usepackage{amsmath}

\title{Assessment of RAG and Fine-Tuning for Industrial Question-Answering-Applications}
\author {
    Jakob Sturm\textsuperscript{\rm 1, 2},
    Josef Pichlmeier\textsuperscript{\rm 1, 3},
    Christian Bernhard\textsuperscript{\rm 1},
    Maka Karalashvili\textsuperscript{\rm 1},
    Johannes Klepsch\textsuperscript{\rm 1},
    Georg Groh\textsuperscript{\rm 2},
    Andre Luckow\textsuperscript{\rm 1, 3}
}
\affiliations {
    \textsuperscript{\rm 1}BMW Group; Munich, Germany\\
    \textsuperscript{\rm 2}Technical University of Munich; Munich, Germany\\
    \textsuperscript{\rm 3}Ludwig Maximilians Universitat; Munich, Germany\\
    jakob.sturm@bmw.de
}

\begin{document}

\maketitle

\begin{abstract}
Large Language Models (LLMs) are increasingly employed in enterprise question-answering (QA) systems, requiring adaptation to domain-specific knowledge.
Among the most prevalent methods for incorporating such knowledge are Retrieval-Augmented Generation (RAG) and fine-tuning (FT).
Yet, from a cost–accuracy trade-off perspective, it remains unclear which approach best suits industry scenarios.
This study examines the impact of RAG and FT on two closed datasets specific to the automotive industry, assessing answer quality and operational costs.
We extend the Cost-of-Pass framework proposed by \citet{Erol2025} to jointly assess output quality, generation cost, and user interaction cost.
Our findings reveal that while premium models perform best out of the box, open-source models can achieve comparable quality when enhanced with RAG. Overall, RAG emerges as the most effective and cost-efficient adaptation method for both closed- and open-source models.
\end{abstract}


\section{Introduction} \label{sec: Introduction}

Large Language Models (LLMs) are critical building blocks for modern AI systems. They are trained on a vast amount of public data and show very strong capabilities in terms of language and knowledge. 
However, this strength is restricted to the boundaries of the training data.
Consequently, many industry applications relying on LLMs underperform expectations \cite{Challapally2025}.
Therefore, adapting LLMs to incorporate organization-specific domain knowledge is a key challenge for their success \cite{Ling2023}.

Retrieval-Augmented Generation (RAG) \cite{lewis2021retrievalaugmentedgenerationknowledgeintensivenlp} and fine-tuning (FT) \cite[e.g.][]{parthasarathy2024ultimateguidefinetuningllms} are widely used strategies for customization. 
RAG retrieves external data at inference time, augmenting the prompt to achieve grounded responses. FT refines model parameters via supervised training on labeled data.
RAG's focus at inference time on relevant knowledge offers great precision; however, this incurs an overhead at runtime. FTs' deep integration enables better generalization, but demands a higher upfront investment. Thus, their choice significantly affects both cost and quality. A combined RAG+FT approach promises the strongest performance through the combination of advantages; however, it also incorporates the associated downsides.

While prior research (see Sec. \ref{sec: RelatedWork}) has compared these methods, typically finding RAG to be more accurate, robust evaluations in industrial settings are lacking.
Moreover, previous studies focus predominantly on accuracy, overlooking holistic cost-performance trade-offs. In practice, stakeholders prioritize these trade-offs: marginal accuracy gains rarely justify drastically higher costs, while budget solutions are rejected if they compromise usability. We aim to contribute to closing these gaps.

To this end, we adapt and extend the Cost-of-Pass framework proposed by \citet{Erol2025}, which incorporates inference cost and performance into a unified expected total cost of solved tasks.
We extend this framework to include human interaction costs, such as verifying LLM outputs (see Sec. \ref{sec: CoP}).
This addition provides a more comprehensive and realistic basis for evaluating customization strategies in industry scenarios. 

From our evaluation, we can identify four main findings:

\textbf{RAG's improved accuracy offsets higher pipeline costs:} Although Retrieval-Augmented Generation is the most expensive architecture to operate, its extended Cost-of-Pass is lowest because it substantially reduces human labor.

\textbf{Premium base models pay off:} When using an off-the-shelf (non-adapted) LLM, the higher-priced model delivers better cost-efficiency than a cheaper alternative.

\textbf{Open-source parity with proprietary giants, when using RAG:} For domain-specific automotive data, small open-source models match the performance of large proprietary models once RAG is applied.

\textbf{Poorly planned deployments can backfire:} An ill-designed setup may cause more cost than having a human complete the task manually, nullifying the benefits of automation.

\section{Datasets} \label{sec: Datasets}
 
To evaluate performance in domain-specific contexts, we utilize two proprietary datasets from the automotive industry provided by the BMW Group: Car User Manual and Vehicle Quality.
These are composed of free-text documents transformed to synthetic QA pairs.
The base data is non-public and partially contains confidential information; therefore, we are unable to release the datasets.
As this data is largely unknown to pre-trained LLMs, it provides robust insights into the generalization and adaptation capabilities of models.

\subsection{Vehicle Quality Dataset} \label{sec: Vehicle_Quality_Dataset}
The Vehicle Quality Dataset encapsulates critical elements of a structured Quality Management (QM) process.
The primary objectives of such QM processes include identifying, documenting, and prioritizing defects, conducting root cause analysis, implementing corrective actions, and institutionalizing best practices to prevent recurrence. 
QM, based on issue tracking and resolution, is a fundamental practice in various domains, including manufacturing and software development.
The Vehicle Quality Dataset is derived from the textual content of the issue tracking tickets, each comprising an initial incident report and a subsequent, more comprehensive, and holistic problem description. We treat each ticket as an individual chunk later on.
The texts are characterized by the use of highly domain-specific terminology, including abbreviations, technical jargon, and alphanumeric identifiers such as part numbers and error codes.
Due to operational time constraints, the writing style often favors brevity and structured, bullet-point formats. Consequently, the textual data departs from conventional natural language, instead presenting as a semi-structured blend of codes, numerical data, and short factual statements.

\subsection{Car User Manual Dataset}
\label{sec: Car_Manual_Dataset}

The basis of the manuals dataset is PDF files covering vehicle-related information and car usage instructions.
The content is structured into fine-grained sections, which we utilize as chunks.
To unify the multi-modal nature of the PDF files, GPT-4o Vision was used to convert the PDFs to text, including descriptions of images and icons.
The information covered in the Car User Manuals text represents a special domain, yet as these manuals are customer-facing, they stick to natural language.
Although these manuals are not freely available, they are publicly accessible to customers. Consequently, we assume that portions of their content may have been included in LLM pre-training.

\subsection{Synthetic QA Pair Generation}\label{sec: qa_gen}

The QA pairs were generated by instruction prompting GPT-4o on the text entities, which correspond to the ticket texts in the case of the Vehicle Quality dataset, and the sections, in the case of the Car User Manual dataset. Furthermore, the LLM was instructed to exhaust each given context. This means that, when possible, the same context serves as the basis for more than one QA pair. 
Due to the technical and often complex nature of the Vehicle Quality tickets, some generated questions were suboptimal. To address this, we incorporated a quality estimation step into the prompting process. Alongside each QA pair, the model returned a quality score ranging from zero (poor) to four (excellent). Only QA pairs receiving the highest score were retained for further use.

\subsection{Dataset statistics}

\newcommand{\rothead}[1]{\rotatebox[origin=c]{90}{\strut #1}}
\begin{table}[t]
    \centering
    \small
    \setlength{\tabcolsep}{4pt}
    \begin{tabular}{l|ccccccc}
        \toprule
        Dataset & $NumC$ & $L_C$ & $L_A$ & $L_Q$ & Odd & \# Train & \# Test \\
        \hline
        Manuals & 2168 & 107 & 19 & 15 & 1\% & 13936 & 1417 \\
        Quality & 3456 & 393 & 25 & 21 & 16\% & 10704 & 2294 \\
        \bottomrule
    \end{tabular}
    \caption{Dataset Statistics (Lengths given in tokens)}
    \label{tab:dataset_statistics}
\end{table}

Table \ref{tab:dataset_statistics} reports basic statistics about the data sets: the number of chunks ($NumC$), the average length of each chunk ($L_C$), the average question length ($L_Q$), and the average length of the reference answers ($L_A$).

To further characterize the linguistic specificity of each dataset, we introduce a metric termed Oddness (Odd). This measure captures the degree to which the wording deviates from what is commonly encountered in the pretraining corpus of an LLM. We operationalize this via an LLM-as-a-Judge approach, using GPT-4o to identify words it deems "unusual" or "odd" within a given context.
The Oddness score is computed as the ratio of the number of identified odd words to the total word count of the text.
The mean Oddness per dataset is given in Table \ref{tab:dataset_statistics}.
The Car User Manual dataset, designed for consumer-level understanding, exhibits a low proportion of uncommon words.
In contrast, the Vehicle Quality dataset, authored by and for domain experts, demonstrates a substantially higher Oddness score, reflecting its specialized language.

We partition the dataset into training and test splits and report the respective amounts in Table~\ref{tab:dataset_statistics}. The training set is used exclusively for FT, while the test sets provide ground-truth question–answer pairs for all pipeline evaluations.

\section{Experimental Setup} \label{sec: ExperimentalSetup}
\subsection{Evaluation scope and fairness}
A perfectly fair comparison between RAG and FT is difficult, as the two approaches incorporate knowledge at different stages. However, utilizing a train–test split provides a symmetric evaluation setting: at test time, both systems answer previously unseen questions, RAG, because it is not trained on questions, and FT, because it is trained only on the training split.
As described above, the training split is drawn from exhaustively generated QA pairs, aiming to approximate coverage of the underlying corpus knowledge. While this approximation is inherently limited, it reflects realistic constraints encountered in real-world applications.
Consequently, FT must generalize from training to test questions, while RAG must derive answers from raw document chunks.
We therefore consider this setup to provide a sufficiently fair basis for comparison.
Questions whose required knowledge is absent from both the training data and the retrieval corpus are considered out of scope.

\subsection{Pipelines}

We evaluate four language models in our analysis: GPT-4o, GPT-4o-mini, LLaMA3.3-70B, and LLaMA3.2-3B. These models represent two major families: proprietary models from a leading commercial vendor (OpenAI) and prominent open-source models (Meta's LLaMA series). The inclusion of both large and smaller variants allows us to assess the impact of model size on performance.
Each model is evaluated under four distinct pipeline configurations:

\textbf{Base}: The model is used without any adaptation.

\textbf{Fine-tuned (FT)}:
Each model is fine-tuned on the respective datasets. GPT models are fine-tuned using Azure OpenAI’s fine-tuning API, with one training epoch to minimize overfitting and default hyperparameters.
LLaMA models are fine-tuned locally using LoRA.
We use a rank of 64, a learning rate of $1 \cdot 10^{-4}$, a batch size of 32 and two epochs. All LLaMA models were fine-tuned and deployed in the bfloat16 format. 

\textbf{Retrieval-Augmented Generation (RAG)}: The model input is supplemented with retrieved contexts: Vehicle Quality tickets or Car User Manual sections.
For retrieval, the proprietary stack uses the text-embedding-ada-002 model, while the open-source setup uses the bge-small embedding model \cite{bge_embedding}.
The number of retrieved samples is set to $k=10$ in the GPT stack and to $k=5$ in the LLaMA stack.
These configurations represent a pragmatic compromise between retrieval coverage and robustness to noise and are based on preliminary experiments.

\textbf{FT + RAG}: This hybrid setup combines the fine-tuned model with RAG-enhanced inputs.

\subsection{Evaluation}
In open-ended text-generation tasks, traditional NLP metrics, such as ROUGE or METEOR \cite{Lin_2004,Banerjee_Lavie_2005}, correlate less with human judgment, since they are not capable of measuring more involved text qualities such as factual correctness, coherence, or relevance \cite{liu-etal-2023-g}.
This limitation is particularly pronounced in our setting, as the task constitutes a genuinely open-ended scenario, which is reflected in the comparatively long average answer lengths observed across both datasets (see Table \ref{tab:dataset_statistics}).
It has been shown that LLM-as-a-Judge effectively addresses this limitation and offers a valid alternative to expensive human evaluation \cite{chiang-lee-2023-large, 10.5555/3666122.3668142, judgelm, liu-etal-2023-g}.
As the presented datasets focus on knowledge-based QA scenarios, we evaluate correctness using LLM-as-a-Judge with GPT-4o.

\subsection{Cost-of-Pass} \label{sec: CoP}
For a holistic analysis, we adopt the Cost-of-Pass concept proposed by \citet{Erol2025}. This framework allows for the unified evaluation of price and quality for LLM-based systems. They conceptualize LLMs as stochastic producers, meaning that lower-capability models require more sampling to generate correct outputs. This makes the trade-off between quality and cost explicit: a less accurate system must be executed more frequently, incurring higher cumulative cost, to achieve the same result.
Thus, the expected Cost of Pass ($CoP$) depends on the expected cost $G$ for a single execution of the LLM pipeline and the expected success rate $S$. This process is modeled by Equation \ref{eq:cost_of_pass_V3}, where the recursion is resolved as a geometric series:
{\small
\begin{align}
    CoP = G + CoP*(1-S) = \frac{G}{S} \label{eq:cost_of_pass_V3}
\end{align}
}%
To extend this model, we propose two enhancements leading to Equations \ref{eq:K_rec_V2} and \ref{eq:K_closed_V2}:
First, we incorporate a validation cost $V$, representing the human effort required to assess whether an output is acceptable or the system needs to be rerun.
Second, we assume users will not rerun the system indefinitely until the correct result appears by chance. Drawing again from \citeauthor{Erol2025}, we introduce the human generation cost $H$ as a fallback, which is incurred if the user gives up on rerunning and manually tackles the task, e.g., by searching for the answer using more traditional, slower methods.
We split the recursive term in Equation \ref{eq:cost_of_pass_V3} into two components: rerunning and human fallback, proportionally weighted by the rerun probability $R$, which captures the user’s willingness to retry. This yields the revised recursive Equation~\eqref{eq:K_rec_V2}:
{\small
\begin{align}
    CoP_{ex} = G + V + (CoP_{ex} * R + H * (1-R))\,(1-S) \label{eq:K_rec_V2}
\end{align}
}%
Solving this recursion gives the closed-form Equation~\eqref{eq:K_closed_V2}:
{\small
\begin{equation}
    CoP_{ex} = \frac{G + V + H\,(1-R)\,(1-S)}{1 - R\,(1 - S)} \label{eq:K_closed_V2}
\end{equation}
}%
Note that this extended Cost-of-Pass  ($CoP_{ex}$) is equivalent to $CoP$ when the repetition rate $R$ is assumed to be one and the validation cost $V$ to be zero.
While this extended Cost-of-Pass model still has limitations, we believe it more accurately reflects the collaborative dynamics between LLMs and human users. 

The cost estimations for $G$ are derived from Azure OpenAI pricing for the GPT-based model stack and from AWS EC2 instance rates for the open-source deployment. Comprehensive cost breakdowns are provided in the Appendix.
$S$ is set to the correctness rate based on LLM-as-a-judge.
We assume a conservative estimate of human labor costs at \$1 per human-generated answer and \$0.10 per validation instance. These values are at the lower end of costs for complex tasks reported by \citet{Erol2025}. 
Additionally, preliminary analysis indicated that increasing labor costs does not affect the relative trends reported below; therefore, these values serve as a consistent illustrative baseline.

\section{Results}
While the Appendix includes detailed information on model accuracy, request costs, and extended Cost-of-Pass, this section focuses on selected experiments that best illustrate the key findings. All system differences discussed in this section are statistically significant (see Appendix for details).

\begin{figure}[t]
\includegraphics[scale=0.65]{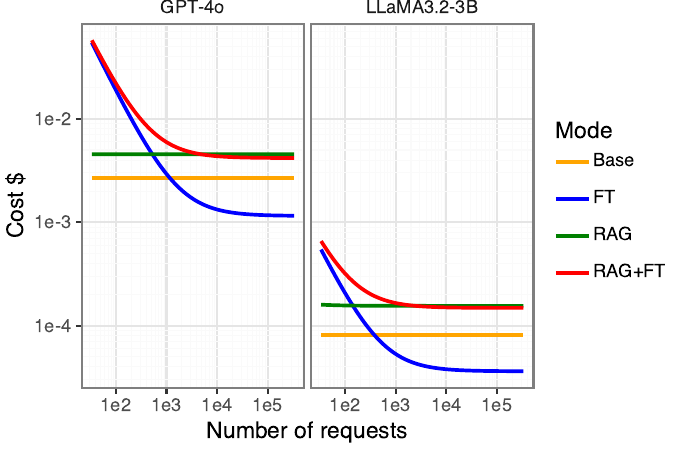}
\caption{Cost over Requests for Manuals Dataset}
\label{fig:per_request_cost_V2}
\end{figure}

As illustrated in Figure~\ref{fig:per_request_cost_V2}, the per-request cost for GPT-4o is consistently higher than that of the LLaMA3.2-3B, irrespective of the chosen pipeline configuration: Base, FT, RAG, or RAG+FT. This cost discrepancy suggests that smaller models may offer a cost-effective alternative.
While this is intuitive, it is not necessarily obvious, as cloud-hosting of open models may incur additional overheads compared to vendor-optimized solutions.
The figure further demonstrates that RAG-based configurations incur higher costs than the Base model, which can be attributed to the larger number of input tokens. In the case of FT models, the per-request cost depends on overall model utilization, as the substantial upfront FT investment must be amortized over time. Interestingly, we observe that FT models tend to produce shorter responses, which can eventually offset the initial investment by reducing token usage per query. 

\begin{figure}[t]
\includegraphics[scale=0.65]{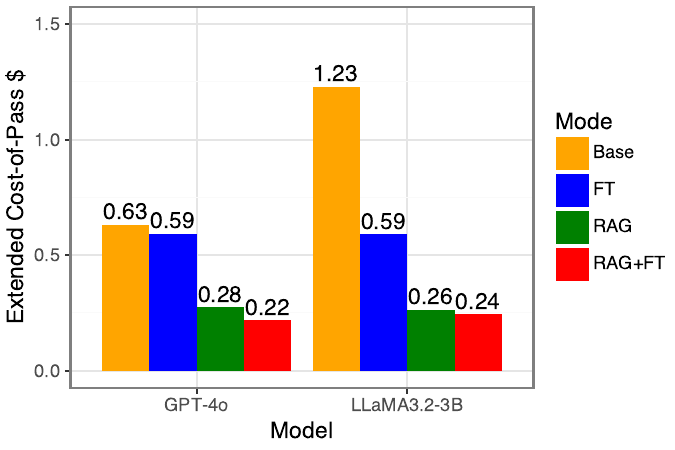}
\caption{Extended Cost-of-Pass for Manuals dataset}
\label{fig:cop_V2_manuals}
\end{figure}

Figure~\ref{fig:cop_V2_manuals} reports the $CoP_{ex}$ results for GPT-4o and LLaMA3.2-3B on the Manuals dataset.
We observe that the Base GPT model demonstrates a lower expected total cost compared to the LLaMA model. This cost advantage is primarily driven by GPT’s higher accuracy on the dataset. Although GPT incurs higher inference costs, its superior performance significantly reduces the need for human validation and intervention, ultimately lowering the overall cumulative cost. Therefore, we conclude that investing in \textbf{the premium model would pay off} in this usage scenario.
Inspecting the effect of RAG reveals its dominating positive impact.
The drastically \textbf{improved accuracy is making up for the increased pipeline costs}, with the primary benefit being a significant reduction in human effort.

\begin{figure}[t]
\includegraphics[scale=0.65]{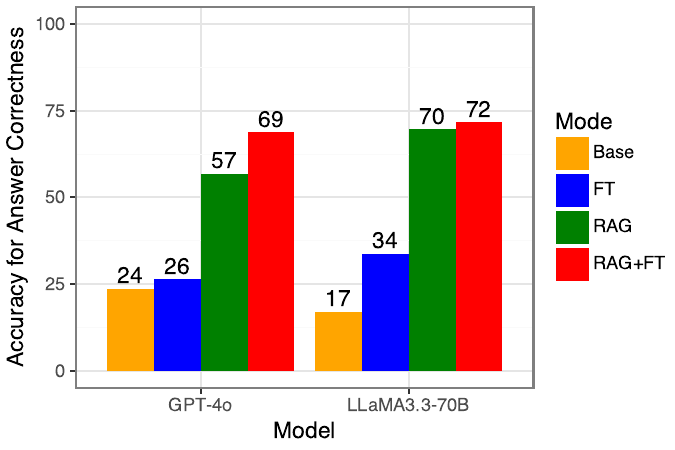}
\caption{Answer correctness for Manuals Dataset
}
\label{fig:V2_accuracy_manuals}
\end{figure}

Figure \ref{fig:V2_accuracy_manuals} compares accuracy values of GPT-4o and LLaMA3.3-70B. The results reveal that, while smaller open-source models encode less prior knowledge, they perform competitively under RAG. \textbf{The open-source model achieves parity} in leveraging contextual information from unfamiliar domains.

\begin{figure}[t]
\includegraphics[scale=0.65]{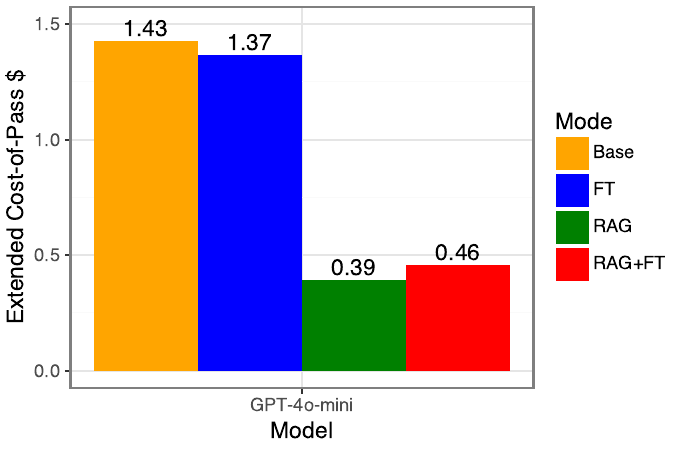}
\caption{Extended Cost-of-Pass for Vehicle Quality dataset}
\label{fig:V2_cop_pmp}
\end{figure}

Finally, we emphasize that \textbf{suboptimal configurations can backfire} substantially. As shown in Figure \ref{fig:V2_cop_pmp}, both the Base and FT models are comparably expensive on the challenging Vehicle Quality dataset. Their low accuracy leads to increased amounts of repetitions, human validations and human interventions. Despite the system's intended cost-saving benefits, this results in higher cumulative expenses (e.g. Base: \$1.43) compared to manual processing (\$1).

\section{Related Work} \label{sec: RelatedWork}

Multiple studies have investigated whether RAG or FT is best suited for injecting knowledge into LLMs \cite{Nguyen_2024, Lakatos_2024, Salemi_2024}. \citet{Ovadia_2023} have shown that RAG consistently outperforms unsupervised FT for both in and out-of-distribution knowledge. Their study examines multiple domains, including anatomy, biology, and physics. A similar conclusion has been made by \citet{Soudani_2024}. The study evaluated the performance of RAG and FT on different sets of Wikipedia articles, categorized by the popularity of the articles. They have shown that while FT can increase performance, RAG is surpassing FT by a significant margin, especially for less popular articles. A more nuanced conclusion has been made by \citet{Balabuer_2024} who evaluated the question RAG vs FT for the domain of agricultural data. They analyzed both approaches in terms of accuracy, cost, and adaptability, showing that while FT improves performance, a hybrid approach combining RAG and FT achieves the best results.
\citet{Wu_2024} evaluated the efficiency of API-based FT for knowledge infusion, showing that commercial FT services struggle to generalize beyond training examples.
In comparison, our work evaluates RAG and FT on industry domain datasets covering real-world use-cases.
Furthermore our structured approach applying the extended Cost-of-Pass allows for a more holistic analysis.

\section{Conclusion} \label{sec: Conclustion}
This study compares RAG and fine-tuning (FT) for adapting LLMs to industrial question answering tasks. We utilize synthetic datasets derived from real-world automotive data, spanning a range from customer-facing texts with few domain-specific terms to expert-oriented content featuring technical language.
We argue that answer accuracy alone is insufficient for evaluating LLM adaptation strategies. Effective comparison must incorporate the generation cost and the required user interaction effort. To this end, we apply an extended Cost-of-Pass evaluation based on the work of \citet{Erol2025}.
Experiments show that premium LLMs offer good value in out-of-the-box usage compared to smaller open-source models: their higher accuracy offsets the increased inference costs. Furthermore, RAG substantially improves answer quality and consistently outperforms FT, reducing the need for human intervention and demonstrating cost efficiency in knowledge-intensive settings.
Interestingly, while premium models excel in Base configurations, their lead diminishes under RAG. In this setup, open-source models achieve comparable performance.
The cost analysis further reveals that LLM-human collaboration can also increase total cost depending on pipeline accuracy. To be cost-effective, the system must reach a performance threshold that reduces the need for human intervention sufficiently.

\appendix

\section{Appendices}

\subsection{Price Analysis} \label{sec: price}
In the context of LLM usage the total costs of one specific method might not be immediately apparent.
Therefore, the next section offers greater transparency into the cost of individual requests across different configurations and scenarios.

\begin{figure*}[t]
  \small
  \noindent
  \setlength{\jot}{10pt}       
  \begin{align}
    G_{\text{Base}} &= (\text{LenQ}+\text{LenP})\,PIT
                      + \text{LenA}_{\text{Base}}\,POT
    \label{eq:generation_cost_base} \\
    G_{\text{RAG}} &= \frac{\text{NumC}\,\text{LenC}\,\text{PET}}{\text{NumRL}}
                     + \text{VDB}
                     + \bigl(K\,\text{LenC} + \text{LenQ} + \text{LenP}_{\text{RAG}}\bigr)PIT
                     + \text{LenA}_{\text{RAG}}\,POT
    \label{eq:generation_cost_rag} \\
    G_{\text{FT}} &= \frac{\text{NumFTQA}\,\text{LenQA}\,\text{PFT}}{\text{NumRL}}
                    + \frac{\text{PH}}{\text{NumRH}}
                    + \bigl(\text{LenQ} + \text{LenP}_{\text{FT}}\bigr)PIT_{\text{FT}}
                    + \text{LenA}_{\text{FT}}\,POT_{\text{FT}}
    \label{eq:generation_cost_ft} \\
    G_{\text{FT+RAG}} &= \frac{\text{NumC}\,\text{LenC}\,\text{PET}}{\text{NumRL}}
                        + \text{VDB}
                        + \frac{\text{NumFTQA}\,\text{LenQA}\,\text{PFT}}{\text{NumRL}}
                        + \frac{\text{PH}}{\text{NumRH}} \notag \\
    &\quad + \bigl(K\,\text{LenC} + \text{LenQ} + \text{LenP}_{\text{RAG}}\bigr)PIT_{\text{FT}}
           + \text{LenA}_{\text{FT+RAG}}\,POT_{\text{FT}}
    \label{eq:generation_cost_ft_rag}
  \end{align}
\end{figure*}

Table \ref{tab:price_comps} provides an overview of the important factors that influence the cost per request along with the used abbreviations.

In the classical scenario, where only a base model is used, the inference costs are determined by processing the input and output tokens (see Equation \eqref{eq:generation_cost_base}).
For RAG and FT, additional cost components must be considered.
With RAG (see Equation \eqref{eq:generation_cost_rag}), there is an extra overhead for generating embeddings from a text corpus and retrieving relevant documents during inference. Furthermore, as we input the retrieved context into the LLM, the number of input tokens increases significantly, thereby increasing the inference costs. 
The number of training samples and tokens in each sample determines the initial training cost for FT. Additionally, hourly serving costs of the deployed fine-tuned model might need to be considered, depending on the provided pricing schema (see Equation \eqref{eq:generation_cost_ft}). When
RAG is used in combination with a fine-tuned model, the respective cost factors accumulate (see Equation \eqref{eq:generation_cost_ft_rag}).
It is important to note that the cost per request decreases for RAG and fine-tuning (FT) as more requests are served from a single system initialization, since the initial fixed costs are distributed across a greater number of requests. 

Preliminary analysis has shown that the lifetime (i.e., when the system is retrained or the knowledge is freshly embedded) has a less pronounced impact compared to other factors such as pipeline type, model, and the number of requests per hour. We therefore choose a single value for all further analysis: a lifetime of seven days. This duration should be sufficiently flexible for most systems, including those with high demands for up-to-date information. By that point, costs have largely stabilized compared to longer lifetimes.

\begin{table}[!htbp]
    \centering
    \begin{tabular}{l|ccccccc}
         Factor & \multicolumn{4}{c}{Model} \\
         & \multicolumn{2}{c}{GPT-4o} & \multicolumn{2}{c}{GPT-4o-mini} \\
         & Base & FT & Base & FT \\
        \hline
        $PIT$   & 2.75 & 2.75 & 0.165 & 0.165 \\
        $POT$   & 11   & 11   & 0.66  & 0.66  \\
        $PFT$   & -    & 27.5 & -     & 3.3   \\
        $PH$    & -    & 1.7  & -     & 1.7   \\
    \end{tabular}
    \caption{Price components for the used GPT Models. $PIT$, $POT$ and $PFT$ given in \$/1mT; $PH$ given in \$/h.}
    \label{tab:price_table_models}
\end{table}

The price calculations for the GPT models are based on the OpenAI cost catalog \cite{azure_openai}. Table \ref{tab:price_table_models} provides an overview for the LLMs. The selected region is Sweden Central; "US/EU – Data Zones" are used for the base models, and "Regional" for fine-tuning. Normal input mode is selected instead of cached input.
For embedding generation, OpenAI's text-embedding-ada-002 model was used, with a price per token (PET) of \$0.10 per 1 million tokens.

\begin{table}[!htbp]
    \centering
    \begin{tabular}{l|ccccccc}
        Factor & \multicolumn{4}{c}{Model} \\
         & \multicolumn{2}{c}{LLaMA3.3-70B} & \multicolumn{2}{c}{LLaMA3.2-3B}\\
         & High & Low & High & Low \\
        \hline
        $PIT$   & 0.38 & 6.45 & 0.1 & 0.52  \\
        $POT$   & 1.07 & 11.97& 0.24& 1.1   \\
        $PF$   & \multicolumn{2}{c}{11.86} & \multicolumn{2}{c}{2.96} \\
    \end{tabular}
    \caption{Price components for the used LLaMA Models. $PIT$ and $POT$ given in \$/1mT; $PF$ given in \$.}
    \label{tab:price_table_models_llama}
\end{table}

We assume that LLaMA, as an open-source model, would be operated on dedicated hardware in the cloud. This deployment strategy mitigates high upfront costs and provides greater transparency in cost distribution over time. The assumed hardware configuration is an AWS EC2 Linux-based p4d.24xlarge instance, currently priced at USD 23.71569 per hour. We selected this instance type due to its integration of the well-established H100 GPU. It is assumed that fine-tuning each model requires no more than one hour on our datasets. The large model utilizes four of the eight available GPUs, while the small model requires only one. This configuration allows for resource sharing across tasks or potential use of smaller instances; hence, the effective cost is scaled down accordingly. The price per finetuning token is derived by distributing the estimated finetuning price over the training dataset.
The cost per token on self-hosted machines depends on hardware utilization, as hardware expenses represent fixed costs that must be amortized over the number of served requests. However, the number of served requests is influenced not only by request volume but also by the system’s capacity for parallel processing. Based on performance evaluations, we estimate the cost within a range that reflects both low- and high-utilization scenarios. The low-utilization scenario assumes a steady load of six concurrent users, whereas the high-utilization scenario assumes operation at maximum throughput.
A summary of the pricing components is provided in Table~\ref{tab:price_table_models_llama}.

We assume the cost of running the vector database and performing the semantic search is negligible, as it does not impose significant resource demands.

The price per request is significantly influenced by the number of tokens used, which is affected by several factors.
On one hand, there are factors tied to the use case and its data: the number of chunks ($NumC$), the average length of each chunk ($LenC$), the average length of questions and questions-answer pairs ($LenQ$ and $LenQA$). Table \ref{tab:dataset_statistics} provides an overview for our use cases. $NumFTQA$ for the Manuals is 13936 and for Vehicle quality 10704.
On the other hand, there are factors that come directly from the pipeline setup: the length of the prompt template used ($LenP$ and $LenP_{RAG}$), and $K$, the number of retrieved chunks in RAG. These are summarized in Table \ref{tab:configs}.
\begin{table}[t]
    \centering
    \begin{tabular}{l|c}
        Factor & Value\\
        \hline
        $K$    & 10 (GPT), 5 (LLaMA)  \\
        $LenP$ & 300 \\
        $LenP_{RAG}$ & 350 \\
    \end{tabular}
    \caption{Implementation-dependent factors influencing the number of tokens used. We define constant values across use cases and models.}
    \label{tab:configs}
\end{table}
The length of the generated answer depends on the processed information, model characteristics, and the prompt used. The observed values are shown in Table \ref{tab:lenA}. It is worth noting that the prompts used did not impose any constraints on answer length.
\begin{table}[t]
    \centering
    \begin{tabular}{l|cccc}
        Model & Pipeline & Manuals & Quality\\
        \hline
        4o & Base    & 166 & 52 \\
        4o & FT      & 26 & 35 \\
        4o & RAG     & 55 & 55 \\
        4o & RAG+FT  & 21 & 45 \\
        4o-mini & Base   & 159 & 65 \\
        4o-mini & FT     & 26 & 21 \\
        4o-mini & RAG    & 59 & 96 \\
        4o-mini & RAG+FT & 28 & 46 \\
    \end{tabular}
    \caption{$LenA$ for GPT models}
    \label{tab:lenA}
\end{table}

\begin{table}[t]
    \centering
    \begin{tabular}{l|cc}
    Model & Pipeline & Manuals \\
    \midrule
    3.2-3b & BASE & 207 \\
    3.2-3b & FT &20 \\
    3.2-3b & RAG & 51 \\
    3.2-3b & RAG+FT & 22 \\
    3.3-70b & BASE & 233 \\
    3.3-70b & FT & 19 \\
    3.3-70b & RAG & 88 \\
    3.3-70b & RAG+FT &  22 \\
    \end{tabular}
    \caption{$LenA$ for llama models}
    \label{tab:lenA_llama}
\end{table}

\begin{figure}[t]
\includegraphics[scale=0.65]{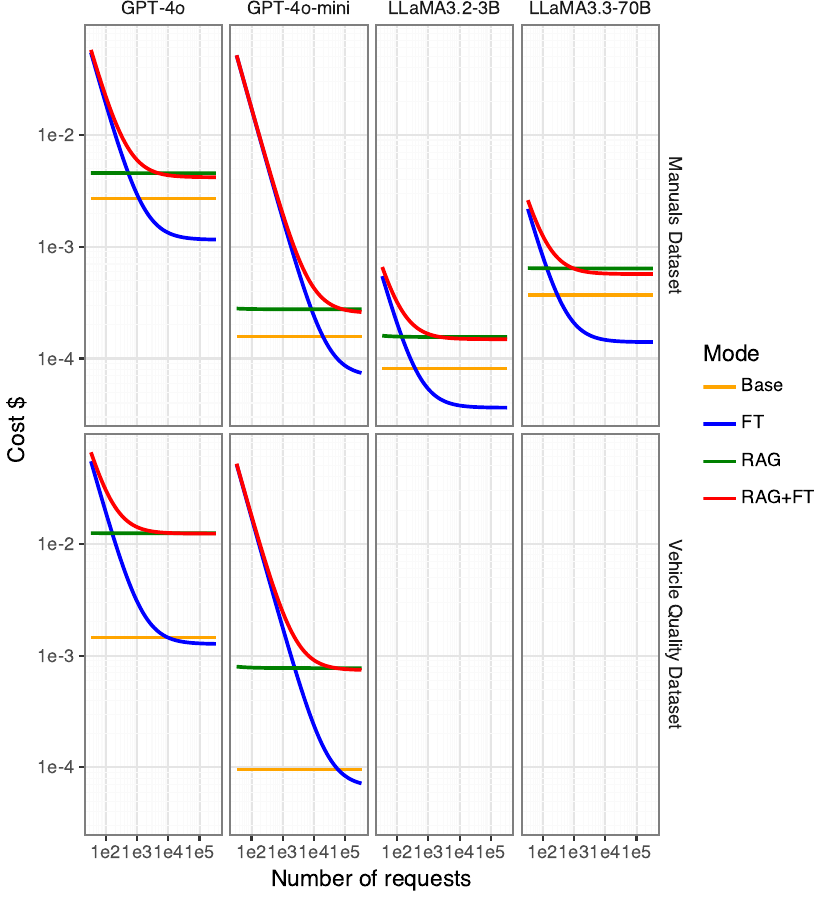}
\caption{Price per request}
\label{fig:per_request_cost}
\end{figure}

\begin{table*}[t]
    \centering
    \begin{tabular}{l|l|ccc}
        Abbreviation & Factor & RAG & FT & Base \\
        \hline
        $NumC$  & Number of Chunks in VecDB & X & & \\
        $LenC$  & Average Chunk length & X & & \\
        $VDB$   & Execution cost of one IR operation (the ANN search) & X & & \\
        $K$     & Amount of retrieved chunks & X & & \\
        $PET$   & Price per Embedded Token & X & & \\
        $PIT$   & Price per Input token & = & X & = \\
        $POT$   & Price per Output Token & = & X & = \\
        $PFT$   & Price for the Fine-Tuning per Token & & X & \\
        $PF$    & Total Price for the Fine-Tuning (used for the LLaMA models) & & X & \\
        $PH$    & Price for Model Hosting per hour & & X & \\
        $NumFTQA$ & Number of QA pairs for the Fine-Tuning & & X & \\
        $LenQA$ & Average Length of the QAs & & X & \\
        $LenP$ & Length of the Prompt template & X & = & = \\
        $LenA$ & Length of the generated Answer & X & X & X \\
        $LenQ$  & Length of the question & = & = & = \\
        $NumRH$   & Amount of Requests per 1 hour & & X & \\
        $NumRL$  & Amount of Requests for the whole system Life time & = & = & = \\
    \end{tabular}
    \caption{Factors influencing the per request generation price. The columns RAG, FT and Base track which factor is relevant for which setup. An X means the system has an individual value for the corresponding factor. An = means the system has the same value for this factor as the other systems marked with an =.}
    \label{tab:price_comps}
\end{table*}

\subsection{Complete Result overview} \label{sec: complete_results}
\begin{figure}[!htbp]
\includegraphics[scale=0.65]{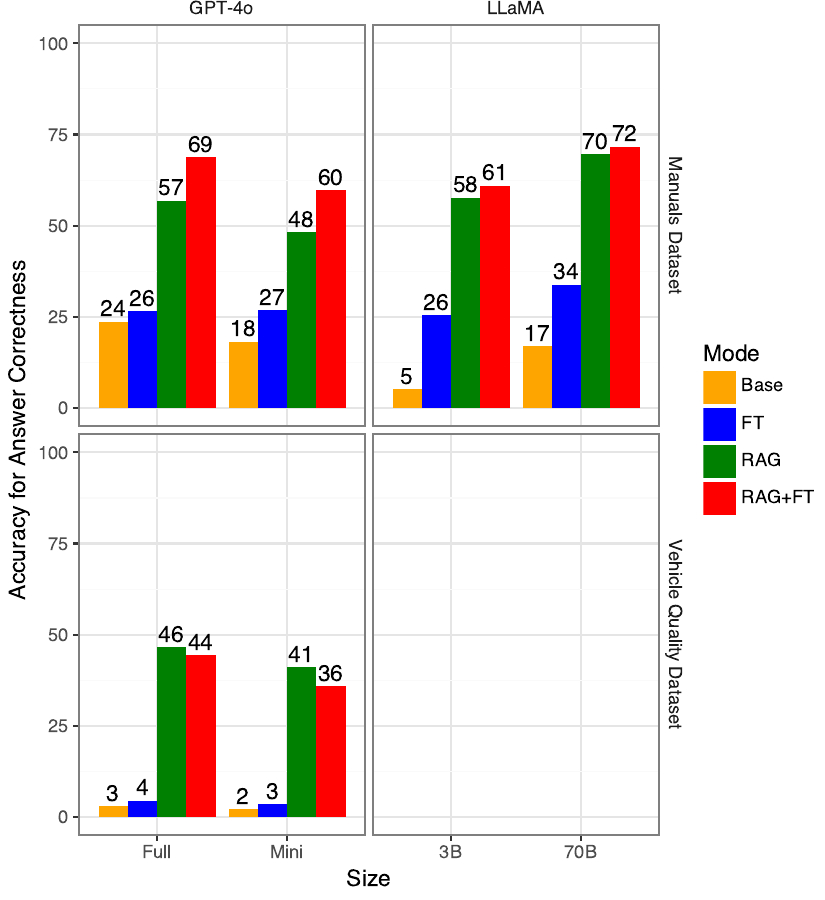}
\caption{Accuracy estimated by LLM-as-a-judge for all experiments}
\label{fig:appendix_accuracy}
\end{figure}

\begin{figure*}[!htbp]
\includegraphics[scale=0.65]{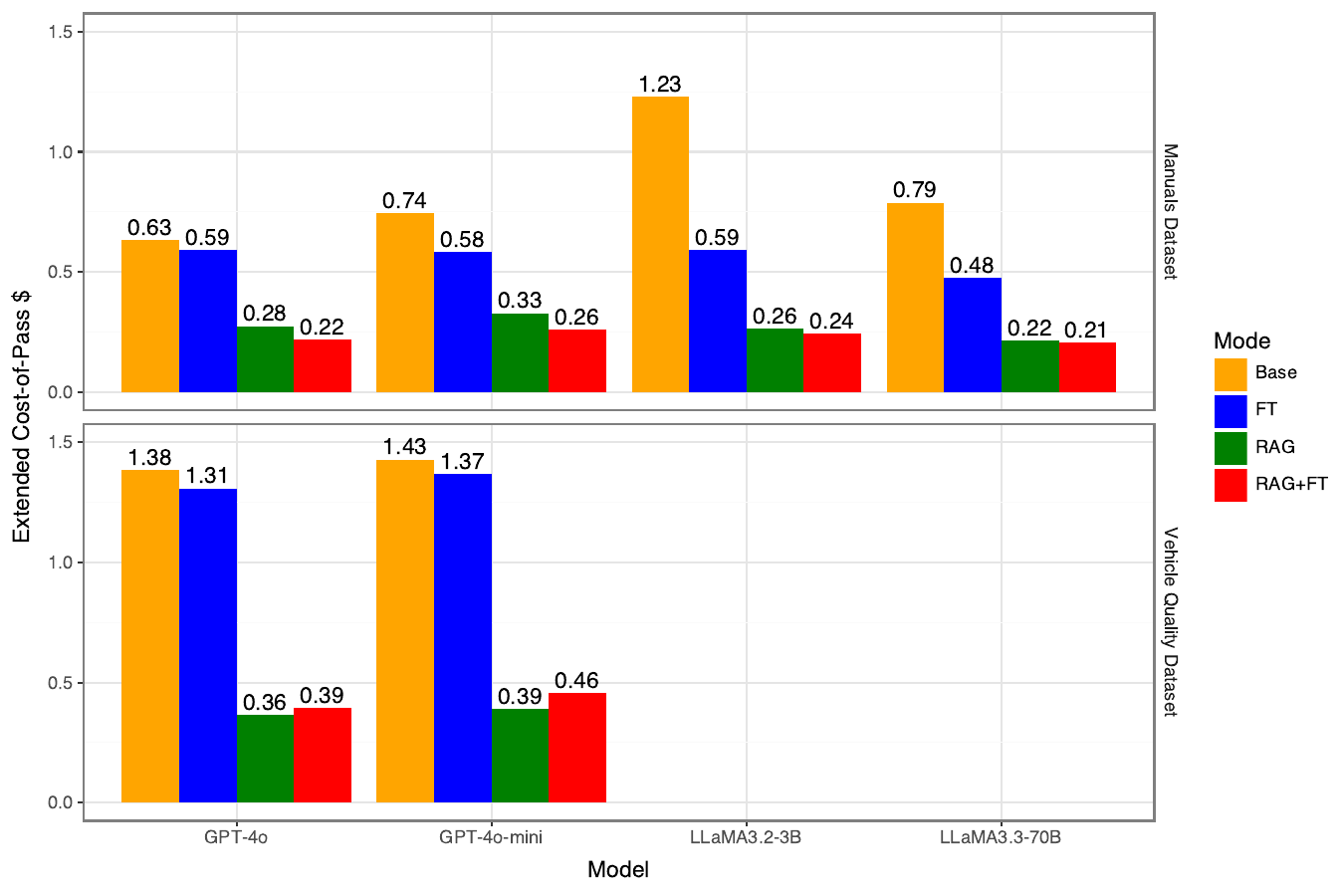}
\caption{Extended Cost-of-Pass for all experiments}
\label{fig:cop}
\end{figure*}

Figure \ref{fig:per_request_cost}, \ref{fig:appendix_accuracy} and \ref{fig:cop} show the holistic overview of the results of all performed experiments.

\subsection{Statistical rigor}
Table \ref{tab:stat_test} reports matrices of multiplicity-adjusted $p$-values testing whether systems differ in performance within each dataset.
We use the \emph{partially overlapping samples $z$-test} (based on the R implementation \texttt{Partiallyoverlapping::Prop.test}), which combines paired and unpaired observations without discarding data.
Each off-diagonal cell is the Holm-adjusted two-sided $p$ for the corresponding row–column contrast; smaller values (e.g., $<\!0.05$) indicate evidence of a difference after correction, larger values indicate no detectable difference.
We observe that systems that exhibit noticeably different accuracies are significantly different.

\begin{table*}[!htbp]
    \centering
    \small
    \setlength{\tabcolsep}{3pt}
    \begin{tabular}{lllllllllllllllll}
        \toprule
         & \rotatebox{90}{\parbox{2.2cm}{\centering GPT-4o}} & \rotatebox{90}{\parbox{2.2cm}{\centering GPT-4o-mini}} & \rotatebox{90}{\parbox{2.2cm}{\centering LLaMA-70B}} & \rotatebox{90}{\parbox{2.2cm}{\centering LLaMA-3B}} & \rotatebox{90}{\parbox{2.2cm}{\centering GPT-4o FT}} & \rotatebox{90}{\parbox{2.2cm}{\centering GPT-4o-mini FT}} & \rotatebox{90}{\parbox{2.2cm}{\centering LLaMA-70B FT}} & \rotatebox{90}{\parbox{2.2cm}{\centering LLaMA-3B FT}} & \rotatebox{90}{\parbox{2.2cm}{\centering GPT-4o RAG}} & \rotatebox{90}{\parbox{2.2cm}{\centering GPT-4o-mini RAG}} & \rotatebox{90}{\parbox{2.2cm}{\centering LLaMA-70B RAG}} & \rotatebox{90}{\parbox{2.2cm}{\centering LLaMA-3B RAG}} & \rotatebox{90}{\parbox{2.2cm}{\centering GPT-4o RAG+FT}} & \rotatebox{90}{\parbox{2.2cm}{\centering GPT-4o-mini RAG+FT}} & \rotatebox{90}{\parbox{2.2cm}{\centering LLaMA-70B RAG+FT}} & \rotatebox{90}{\parbox{2.2cm}{\centering LLaMA-3B RAG+FT}} \\
        \midrule
        GPT-4o &  & 0.000 & 0.000 & 0.000 & 0.599 & 0.387 & 0.000 & 1.000 & 0.000 & 0.000 & 0.000 & 0.000 & 0.000 & 0.000 & 0.000 & 0.000 \\
        GPT-4o-mini & 0.000 &  & 1.000 & 0.000 & 0.000 & 0.000 & 0.000 & 0.000 & 0.000 & 0.000 & 0.000 & 0.000 & 0.000 & 0.000 & 0.000 & 0.000 \\
        LLaMA-70B & 0.000 & 1.000 &  & 0.000 & 0.000 & 0.000 & 0.000 & 0.000 & 0.000 & 0.000 & 0.000 & 0.000 & 0.000 & 0.000 & 0.000 & 0.000 \\
        LLaMA-3B & 0.000 & 0.000 & 0.000 &  & 0.000 & 0.000 & 0.000 & 0.000 & 0.000 & 0.000 & 0.000 & 0.000 & 0.000 & 0.000 & 0.000 & 0.000 \\
        GPT-4o FT & 0.599 & 0.000 & 0.000 & 0.000 &  & 1.000 & 0.000 & 1.000 & 0.000 & 0.000 & 0.000 & 0.000 & 0.000 & 0.000 & 0.000 & 0.000 \\
        GPT-4o-mini FT & 0.387 & 0.000 & 0.000 & 0.000 & 1.000 &  & 0.000 & 1.000 & 0.000 & 0.000 & 0.000 & 0.000 & 0.000 & 0.000 & 0.000 & 0.000 \\
        LLaMA-70B FT & 0.000 & 0.000 & 0.000 & 0.000 & 0.000 & 0.000 &  & 0.000 & 0.000 & 0.000 & 0.000 & 0.000 & 0.000 & 0.000 & 0.000 & 0.000 \\
        LLaMA-3B FT & 1.000 & 0.000 & 0.000 & 0.000 & 1.000 & 1.000 & 0.000 &  & 0.000 & 0.000 & 0.000 & 0.000 & 0.000 & 0.000 & 0.000 & 0.000 \\
        GPT-4o RAG & 0.000 & 0.000 & 0.000 & 0.000 & 0.000 & 0.000 & 0.000 & 0.000 &  & 0.000 & 0.000 & 1.000 & 0.000 & 0.632 & 0.000 & 0.215 \\
        GPT-4o-mini RAG & 0.000 & 0.000 & 0.000 & 0.000 & 0.000 & 0.000 & 0.000 & 0.000 & 0.000 &  & 0.000 & 0.000 & 0.000 & 0.000 & 0.000 & 0.000 \\
        LLaMA-70B RAG & 0.000 & 0.000 & 0.000 & 0.000 & 0.000 & 0.000 & 0.000 & 0.000 & 0.000 & 0.000 &  & 0.000 & 1.000 & 0.000 & 1.000 & 0.000 \\
        LLaMA-3B RAG & 0.000 & 0.000 & 0.000 & 0.000 & 0.000 & 0.000 & 0.000 & 0.000 & 1.000 & 0.000 & 0.000 &  & 0.000 & 1.000 & 0.000 & 0.193 \\
        GPT-4o RAG+FT & 0.000 & 0.000 & 0.000 & 0.000 & 0.000 & 0.000 & 0.000 & 0.000 & 0.000 & 0.000 & 1.000 & 0.000 &  & 0.000 & 0.682 & 0.000 \\
        GPT-4o-mini RAG+FT & 0.000 & 0.000 & 0.000 & 0.000 & 0.000 & 0.000 & 0.000 & 0.000 & 0.632 & 0.000 & 0.000 & 1.000 & 0.000 &  & 0.000 & 1.000 \\
        LLaMA-70B RAG+FT & 0.000 & 0.000 & 0.000 & 0.000 & 0.000 & 0.000 & 0.000 & 0.000 & 0.000 & 0.000 & 1.000 & 0.000 & 0.682 & 0.000 &  & 0.000 \\
        LLaMA-3B RAG+FT & 0.000 & 0.000 & 0.000 & 0.000 & 0.000 & 0.000 & 0.000 & 0.000 & 0.215 & 0.000 & 0.000 & 0.193 & 0.000 & 1.000 & 0.000 &  \\
        \bottomrule
    \end{tabular}
    \hspace{0.5cm}
    
    \begin{tabular}{lllllllll}
        \toprule
         & \rotatebox{90}{\parbox{2.2cm}{\centering GPT-4o}} & \rotatebox{90}{\parbox{2.2cm}{\centering GPT-4o-mini}} & \rotatebox{90}{\parbox{2.2cm}{\centering GPT-4o FT}} & \rotatebox{90}{\parbox{2.2cm}{\centering GPT-4o-mini FT}} & \rotatebox{90}{\parbox{2.2cm}{\centering GPT-4o RAG}} & \rotatebox{90}{\parbox{2.2cm}{\centering GPT-4o-mini RAG}} & \rotatebox{90}{\parbox{2.2cm}{\centering GPT-4o RAG+FT}} & \rotatebox{90}{\parbox{2.2cm}{\centering GPT-4o-mini RAG+FT}} \\
        \midrule
        GPT-4o &  & 0.634 & 0.014 & 1.000 & 0.000 & 0.000 & 0.000 & 0.000 \\
        GPT-4o-mini & 0.634 &  & 0.000 & 0.072 & 0.000 & 0.000 & 0.000 & 0.000 \\
        GPT-4o FT & 0.014 & 0.000 &  & 0.193 & 0.000 & 0.000 & 0.000 & 0.000 \\
        GPT-4o-mini FT & 1.000 & 0.072 & 0.193 &  & 0.000 & 0.000 & 0.000 & 0.000 \\
        GPT-4o RAG & 0.000 & 0.000 & 0.000 & 0.000 &  & 0.000 & 0.500 & 0.000 \\
        GPT-4o-mini RAG & 0.000 & 0.000 & 0.000 & 0.000 & 0.000 &  & 0.073 & 0.000 \\
        GPT-4o RAG+FT & 0.000 & 0.000 & 0.000 & 0.000 & 0.500 & 0.073 &  & 0.000 \\
        GPT-4o-mini RAG+FT & 0.000 & 0.000 & 0.000 & 0.000 & 0.000 & 0.000 & 0.000 &  \\
        \bottomrule
    \end{tabular}
    \caption{
        Pairwise matrices of multiplicity-adjusted $p$-values (Holm–Bonferroni; controls FWER) testing whether systems differ in performance within each dataset (top: Manuals; bottom: Quality). ($p$-values rounded to three decimals; 0.000 denotes $<\!0.0005$.)}
    \label{tab:stat_test}
\end{table*}

\bibliography{AnonymousSubmission/LaTeX/references}

\end{document}